\documentclass[final,3p,times,nopreprintline]{elsarticle}
\usepackage{amssymb}
\usepackage{multirow}
\usepackage{tabularx}
\usepackage{pifont}
\usepackage{amsmath}
\usepackage{setspace}
\usepackage{fancyhdr,graphicx}
\usepackage[ruled,vlined]{algorithm2e}
\SetKw{KwBy}{by}
\usepackage{algpseudocode}
\usepackage{xcolor}

\SetCommentSty{mycommfont}
\usepackage{booktabs}
\usepackage{siunitx}
\usepackage{subcaption}
\usepackage{array}
\usepackage{cuted}

\journal{Pattern Recognition}

\begin{document}

\begin{frontmatter}

\title{STATIC: Surface-normal-guided Two-branch Alignment and Temporal Integration for Consistent Video Depth Estimation}

\author[inst1]{Sunghun Yang}
\author[inst1]{Minhyeok Lee}
\author[inst2]{Suhwan Cho}
\author[inst1]{Jungho Lee}
\author[inst1]{Sangyoun Lee\corref{cor1}}
\cortext[cor1]{First author. Email: sunghunyang98@yonsei.ac.kr}

\affiliation[inst1]{
	organization={School of Electrical and Electronic Engineering, Yonsei University},
	city={Seoul},
	country={South Korea}
}
\affiliation[inst2]{
	organization={GenGenAI},
	city={Seoul},
	country={South Korea}
}

\begin{abstract}
    Video depth estimation is essential for applications such as autonomous driving, AR/VR, and robotics. Recent transformer-based single-image monocular depth estimation models perform well on a single image but struggle with depth consistency across video frames. Traditional methods aim to improve temporal consistency using multi-frame temporal modules or explicit priors like optical flow and camera parameters. However, these approaches face issues such as high memory use, reduced performance with dynamic or irregular motion, and a heavy reliance on the accuracy of external motion estimators. 
    We propose STATIC, a novel model that independently learns temporal consistency in geometrically stable and unstable regions using only two adjacent frames. 
    By leveraging intrinsic geometric cues rather than external motion priors, STATIC generates a depth-derived geometry-aware difference mask from surface normals to identify regions with inter-frame geometric variation by measuring directional variance. 
    For geometrically stable regions, the Masked Stable (MS) module enhances temporal consistency by focusing on structurally stable regions. 
    For geometrically unstable regions, the Surface Normal Similarity (SNS) module models inter-frame correspondences and ensures temporal consistency by measuring feature similarities between frames. 
    A final refinement integrates the independently learned stable and unstable features, enabling STATIC to improve consistent depth prediction across consecutive frames.
    Experiments on KITTI, NYUv2, and Sintel demonstrate strong depth accuracy and temporal consistency, as well as stable performance across the tested frame-rate and camera-motion conditions.
    
\end{abstract}

%%Graphical abstract
% \begin{graphicalabstract}
% \includegraphics[width=\linewidth]{image/fig2/fig2.pdf}
% \end{graphicalabstract}

%%Research highlights
% \begin{highlights}
%     \item A two-frame framework is proposed for temporally consistent video depth estimation.
%     \item Surface-normal variation provides depth-derived, geometry-aware temporal guidance.
%     \item Complementary MS and SNS modules improve consistency in stable and unstable regions.
% \end{highlights}

\begin{keyword}
	Video Depth Estimation \sep Temporal Consistency \sep Surface Normals
\end{keyword}

\end{frontmatter}

%% main text
\section{Introduction}

\begin{figure*}[t]
	\begin{center}
		\includegraphics[width=1.0\linewidth]{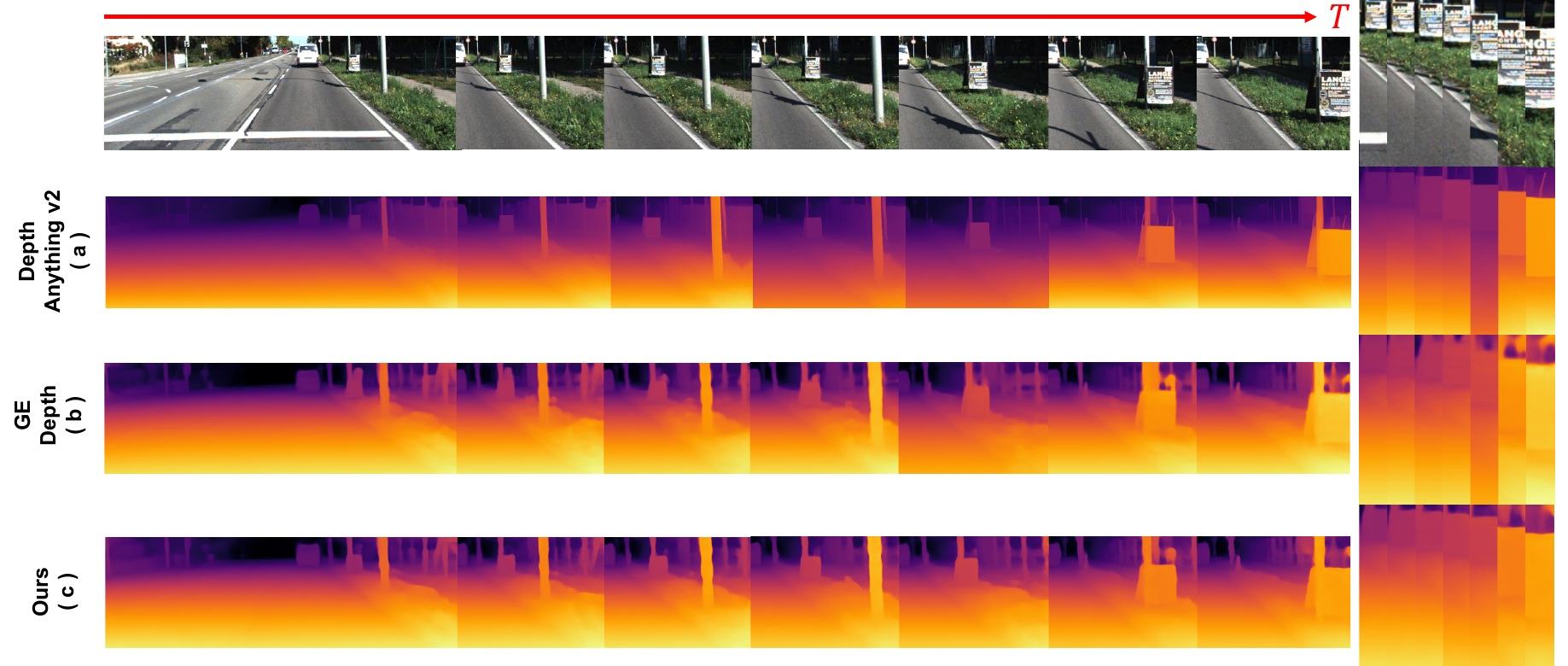}
		\caption{Results of the (a) Depth Anything V2~\cite{yang2024depth}, (b) GEDepth~\cite{yang2023gedepth} and proposed (c) STATIC from sequential frames. Temporal profiles are generated from the marked slice regions to visualize frame-to-frame depth continuity. (c) shows improved ground continuity over the single-frame approach (a) and more stable depth across frames than the compared video depth estimation method (b). For visual comparison, each method is rescaled.}
		\label{fig:fig1}
	\end{center}
\end{figure*}

Depth estimation aims to generate a dense, pixel-level depth map from an RGB image, which is essential in applications such as autonomous driving, AR/VR, and robotics. Recently, transformer-based monocular depth estimation models~\cite{ranftl2021vision, zou2025ssrdepth, tang2024catnet} have demonstrated superior performance due to their robust generalization abilities, relying on large-scale datasets of paired single-view images and depth maps. In many applications, depth maps are typically required from consecutive video frames. Monocular depth estimation can process video frames by predicting depth for each frame individually. However, it cannot consider inter-frame depth relationships. As illustrated in Figure~\ref{fig:fig1} (a), per-frame predictions exhibit low inter-frame consistency. To handle these issues, several recent methods address temporal inconsistency by leveraging the global context inferred from multiple frames, enhancing connectivity across the temporal dimension~\cite{zhang2019exploiting, patil2020don, yasarla2023mamo}. By using multiple-frame inputs, the model improves inter-frame consistency by learning from diverse scenes. Others use explicit cues like optical flow~\cite{xie2020video, wang2021temporal, xiang2023self} or camera parameters~\cite{yang2023gedepth, teed2018deepv2d} with high-quality motion estimation to enhance temporal consistency in depth predictions.

However, each previous temporal consistency method faces major problems. Firstly, many multi-frame methods often focus on broad frame-to-frame changes rather than detailed local motions, making it harder for the model to capture subtle movements. Additionally, processing multiple frames simultaneously leads to inefficiencies and high memory consumption. Secondly, methods that rely on explicit motion cues often become unreliable in the presence of dynamic or irregular motion, which can lead to erroneous temporal propagation. In addition, they typically apply a shared motion representation across the scene, making it difficult to treat moving foreground regions and relatively stable background regions in a differentiated manner. As shown in Figure~\ref{fig:fig1}(b), this limitation results in inconsistent preservation of both foreground objects and background regions across frames. Furthermore, as mentioned in~\cite{lee2024video}, methods can become overly dependent on these additional cues, propagating motion errors directly into the depth predictions.

To address these issues, we propose a novel video depth estimation model, Surface-normal-guided Two-branch Alignment and Temporal Integration for Consistent Video Depth Estimation called STATIC. By using only two adjacent frames, our model avoids the heavy overhead of multi-frame fusion and keeps the temporal input setting lightweight. STATIC constructs a depth-derived geometry-aware difference mask from surface-normal variation and separates geometrically stable and unstable regions for region-specific temporal processing, without relying on external motion priors at inference. This enables the model to handle different types of inter-frame depth variation without explicitly estimating
camera-object motion.

To obtain this guidance map, we use surface normals derived from depth-related representations. Since surface normals encode local surface orientation and shape, they provide an internal geometric cue for identifying regions where temporal depth prediction is likely to become unstable. Thus, the difference mask is computed from inter-frame directional variation rather than inferred from external motion cues.

After separating regions with the difference mask, STATIC processes geometrically stable and unstable regions through two specialized modules. The Masked Stable (MS) module focuses on regions with small inter-frame surface-normal variation, while the Surface Normal Similarity (SNS) module estimates inter-frame correspondences in regions with relatively large geometric variation using depth-normal feature similarity. 
Finally, a refinement process unifies the independently learned stable and unstable features, enabling STATIC to improve temporal consistency across consecutive frames. As shown in Figure ~\ref{fig:fig1}(c), the proposed STATIC improves spatial and temporal consistency in both geometrically stable and unstable regions. Our method was evaluated on widely used datasets: KITTI Eigen split~\cite{geiger2012we}, NYUv2~\cite{silberman2012indoor}
and MPI-Sintel~\cite{butler2012naturalistic}. STATIC achieves state-of-the-art performance across these datasets without relying on external motion priors at inference, using only two adjacent RGB frames. Recent video depth stabilization, foundation-model-based, and diffusion-based approaches address video depth under substantially different supervision settings; we discuss their relationship to our setting in Section 4.4.

\section{Related Work}

%-------------------------------------------------------------------------
\subsection{Video Depth Estimation}

Video depth estimation extends monocular depth prediction by incorporating temporal consistency and motion cues across frames, improving stability and accuracy over time. Unlike single-frame approaches, it captures temporal dependencies to reduce frame-to-frame inconsistencies in dynamic scenes. Recent methods leverage RNNs~\cite{eom2019temporally, patil2020don} and optical flow~\cite{huang2022flowformer} for temporal alignment. RNNs link depth predictions across frames through memory, while optical flow provides explicit motion cues for alignment. Transformers have also been adopted to model both spatial and temporal dependencies. More recently, diffusion models~\cite{patni2024ecodepth, fu2024geowizard} have been explored, mainly improving data diversity rather than directly addressing video depth consistency. Despite this progress, recurrent or multi-frame fusion methods often require high memory and computation, while flow-based methods can propagate motion errors into depth predictions. Our approach addresses these limitations by using only two adjacent frames and intrinsic geometric cues to improve temporal consistency with a lightweight temporal input setting.

%-------------------------------------------------------------------------
\subsection{Surface Normal-based Approaches}

Previous methods~\cite{fu2023multi, ren2023surface} have exploited surface normals to enhance the geometric quality of depth by preserving structures such as corners, boundaries, and planes. These approaches can be broadly grouped into depth-normal joint learning, which jointly estimates depth and normals under mutual geometric constraints~\cite{fu2023multi}, and normal-guided depth refinement, which uses estimated normals to regularize or complete depth structures~\cite{ren2023surface}. In both directions, surface normals serve as per-frame auxiliary geometric constraints for improving single-image depth structure, without requiring explicit camera pose or geometric reconstruction. In contrast to these per-frame uses, we exploit inter-frame surface-normal variation as a cue for temporal region separation, which enables the proposed two-branch design for video depth consistency.

\section{Proposed Approach}

\subsection{Preliminaries}

Surface normals describe the local orientation of scene surfaces and provide a geometrically informative cue closely related to depth. 
Compared with RGB values, which are sensitive to illumination changes, the depth-gradient-based normal representation captures local variations in depth structure rather than image appearance. Although this representation can be affected by depth scale and camera motion, its variation across consecutive frames provides a useful geometric cue for identifying regions with inter-frame changes. In contrast, when the observed surface changes due to object movement, occlusion, or large viewpoint changes, the pixel-wise normal map may exhibit noticeable directional variation.

In our approach, this property is used not to estimate motion explicitly, but to compute inter-frame differences through surface-normal variation. Such variation indicates reduced local geometric consistency and suggests regions where different temporal processing may be beneficial. In this work, stable and unstable regions denote regions with small and large inter-frame geometric variation, respectively. The proposed mask does not aim to recover exact motion, but to locate regions where temporal depth consistency is likely to be unreliable. We therefore use surface-normal variation as a depth-derived geometric cue to guide region-specific temporal processing.

\begin{figure*}[t]
	\setlength{\belowcaptionskip}{-8pt}
	\begin{center}
		\includegraphics[
            width=1.0\linewidth,
            height=0.27\textheight
        ]{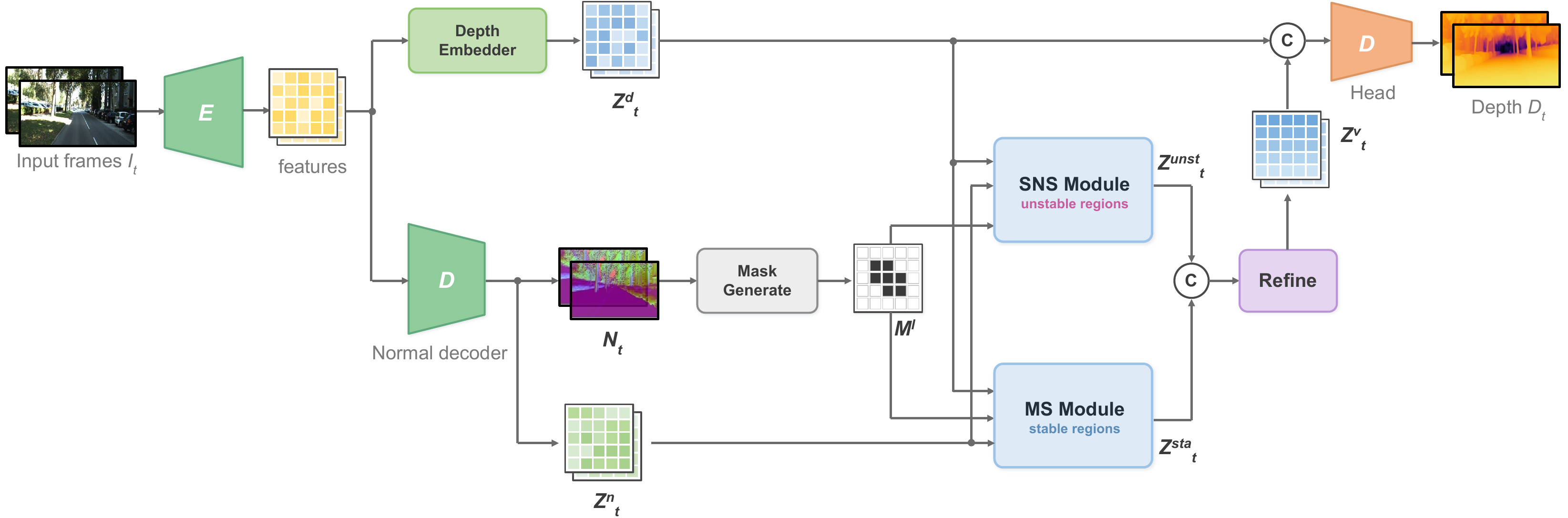}
		\caption{Overall architecture of the proposed STATIC model. The model primarily consists of an encoder, depth embedder, and SNS/MS modules, with a surface normal decoder and head as submodules.}
		\label{fig:fig2}
	\end{center}
\end{figure*}

\subsection{Overall Architecture}

Figure~\ref{fig:fig2} shows the overall architecture of STATIC. Our model uses only two frames, $I_t \in \mathbb{R}^{3 \times H \times W}$, where $t \in \{0, 1\}$, as input.

The encoder, depth embedder, and surface-normal decoder share weights across the two input frames, and the same module weights are applied to each frame. Each frame is passed through the encoder, and the resulting features are used by both the depth embedder and the surface-normal decoder. The depth embedder processes these features to produce the depth context feature $Z^d_t \in \mathbb{R}^{C^d \times \frac{H}{8} \times \frac{W}{8}}$. The surface-normal decoder generates both the surface normal image $N_t \in \mathbb{R}^{3 \times \frac{H}{8} \times \frac{W}{8}}$ and the surface normal feature $Z^n_t \in \mathbb{R}^{C^n \times \frac{H}{8} \times \frac{W}{8}}$.

The predicted surface normal image $N_t$ is used to generate a difference mask, which separates regions according to inter-frame geometric variation. STATIC obtains temporal consistency by employing the Masked Stable (MS) module and the Surface Normal Similarity (SNS) module for geometrically stable and unstable regions, respectively. The outputs of the two modules are then refined and integrated into a video feature, which is concatenated with the depth context feature and upsampled by the depth head to generate consecutive depth maps for both frames.

\begin{figure}[t]
	\setlength{\belowcaptionskip}{-5pt}
	\begin{center}
		\includegraphics[width=1.0\linewidth]{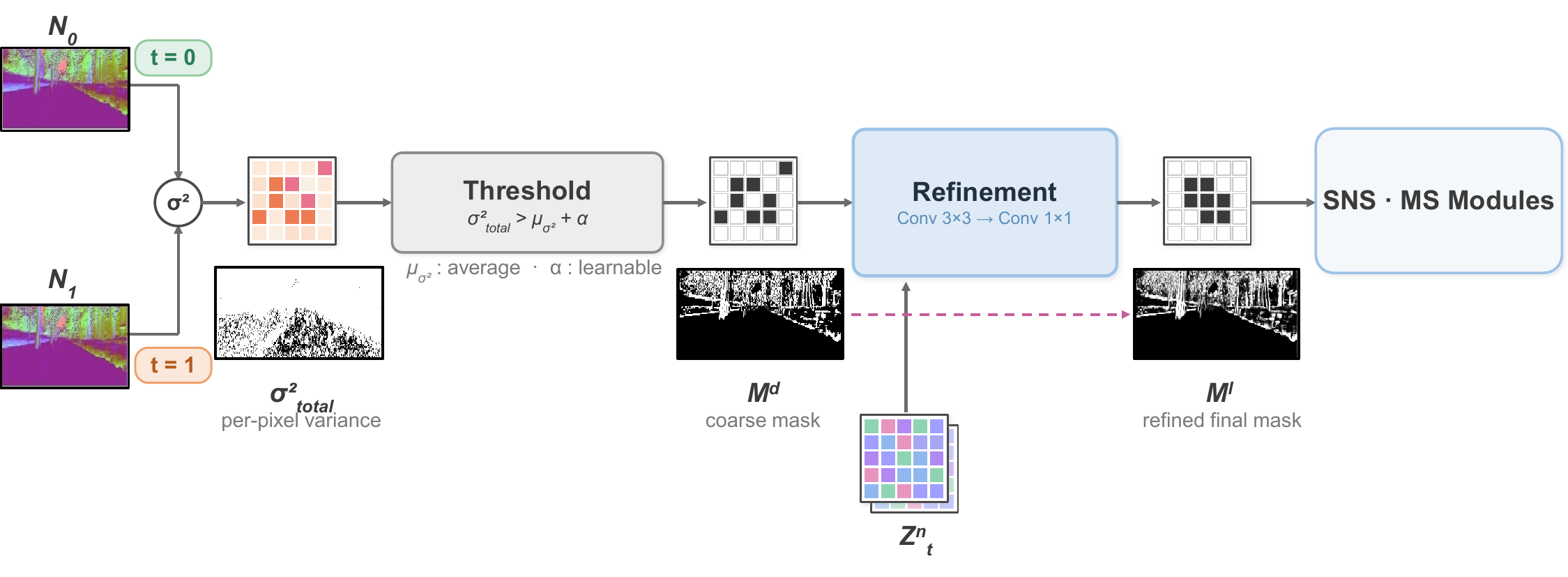}
		\caption{Generation of the proposed difference mask. The pixel-wise directional variance between two surface-normal maps is thresholded to produce the coarse mask $M^d$. The refinement module combines $M^d$ with the surface-normal feature $Z_t^n$ to generate the final mask $M^l$ used by the SNS and MS modules.}
		\label{fig:fig3}
	\end{center}
\end{figure}

\subsection{Difference Mask}

Figure~\ref{fig:fig3} shows the process of generating the coarse difference mask 
$M^d \in \mathbb{R}^{1 \times \frac{H}{8} \times \frac{W}{8}}$. 
To generate $M^d$, we use the surface normal maps produced by the surface normal decoder. 
Surface normals represent local surface orientation, and their directional changes provide a useful cue for identifying inter-frame geometric variation.

Given two adjacent frames, let $\mathbf{N}_t(p) \in \mathbb{R}^{3}$ and 
$\mathbf{N}_{t+1}(p) \in \mathbb{R}^{3}$ denote the surface-normal vectors at pixel location $p$ in frames $t$ and $t+1$, respectively. 
We compute the directional variance in a pixel-wise manner. 
For each axis $k \in \{x,y,z\}$, the mean normal component at pixel $p$ is defined as

\begin{equation}
\begin{aligned}
\mu_k(p) &= \frac{1}{2} \left( N_t^k(p)+N_{t+1}^k(p) \right), \\
\sigma_{\mathrm{total}}^2(p) &= \frac{1}{2} \sum_{k\in\{x,y,z\}} \sum_{\tau\in\{t,t+1\}} \left( N_\tau^k(p)-\mu_k(p) \right)^2 .
\end{aligned}
\label{eq:normal_variance}
\end{equation}

Here, $\sigma^2_{\mathrm{total}}(p)$ is computed independently at each pixel location $p$, rather than over a local spatial neighborhood or the entire image.

Secondly, we compute the mean value $\mu_{\sigma^2}$ of the pixel-wise directional variance map $\sigma^2_{\mathrm{total}}$. 
Since camera movement can affect broad regions, $\sigma^2_{\mathrm{total}}$ may be non-zero even in geometrically stable regions. 
Therefore, we use $\mu_{\sigma^2}$ as a global offset to compensate for this overall camera-induced variation.
Furthermore, we use the learnable parameter $\alpha$ for a more adaptive threshold. 
This process is expressed as follows:
\begin{equation} 
\begin{aligned}
     M^d(p) = 
    \begin{cases} 
    1, & \text{if } \sigma^2_{\mathrm{total}}(p) > \mu_{\sigma^2} + \alpha, \\ 
    0, & \text{otherwise}.
    \end{cases} 
\end{aligned}
\label{eq:coarse_mask}
\end{equation}

Finally, to obtain a clearer mask boundary, we use $M^d$ as a pseudo-label for refinement. 
The refinement module takes the surface-normal feature $Z_t^n$ and the coarse mask $M^d$ as inputs and outputs the refined difference mask $M^l$.
Figure~\ref{fig:fig3} shows the final output mask 
$M^l \in \mathbb{R}^{1 \times \frac{H}{8} \times \frac{W}{8}}$.

\begin{figure}[t]
	\setlength{\belowcaptionskip}{-15pt}
	\begin{center}
		\includegraphics[width=1.0\linewidth]{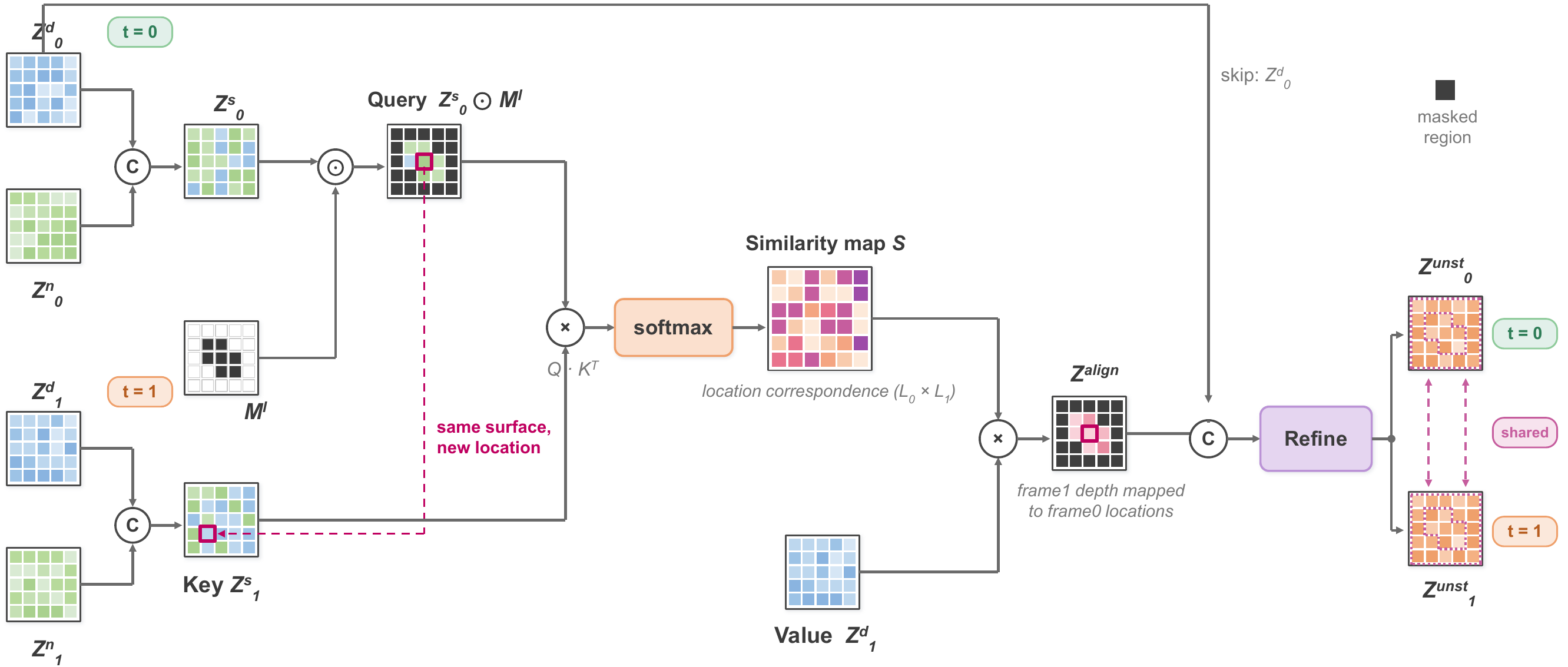}
		\caption{The structure of the SNS module. First, a similarity map $S$ is generated using two features. Then, by multiplying this map with the depth features of other frame, a process similar to warping is performed.}
		\label{fig:fig4}
	\end{center}
\end{figure}

\subsection{Surface Normal Similarity Module}

The SNS module ensures temporal consistency in geometrically unstable regions indicated by $M^l$. Since these regions exhibit large inter-frame geometric variation and low inter-frame correlation, temporal consistency cannot be learned by simply relating features at the same spatial locations. Therefore, the SNS module estimates corresponding locations between the two frames using a similarity map.

As shown in Figure~\ref{fig:fig4}, the detailed structure of the SNS module is presented. First, the depth context feature $Z^d_t$ and surface normal feature $Z^n_t$ are concatenated along the channel axis to form $Z_t^s \in \mathbb{R}^{L_t \times (C_t^d + C_t^n)}$, where $L_t$ represents the total number of pixels $\frac{H}{8} \times \frac{W}{8}$ for each frame. In this joint feature, the depth feature encodes distance similarity and the surface normal feature encodes geometric similarity, so that correspondences are determined by both cues jointly.

Next, $Z_0^s$ multiplied by $M^l$ is used as the query feature, and $Z_1^s$ as the key feature. After performing the dot product between the query and key features, softmax is applied to obtain the location similarity map $S \in \mathbb{R}^{L_0 \times L_1}$. Consequently, in regions of $M^l$ with large geometric variation, the similarity map identifies where the corresponding surface appears in the other frame. This process is expressed as follows:
\begin{equation}
    S = \text{softmax}\left((Z_0^s \odot M^l) \cdot (Z_1^s)^\top\right),
\end{equation}
where $Z_0^s \in \mathbb{R}^{L_0 \times (C_0^d + C_0^n)}$ and $(Z_1^s)^\top \in \mathbb{R}^{(C_1^d + C_1^n) \times L_1}$.

By using the depth context feature $Z_1^d \in \mathbb{R}^{L_1 \times C_1^d}$ as the value feature with $S$, the aligned feature $Z^{align} \in \mathbb{R}^{L_0 \times C_1^d}$ can be obtained. We additionally apply $M^l$ to the aligned feature to retain the attention output only in geometrically unstable regions. This maps the next-frame depth context to the corresponding locations in the current frame, acting as a warping-like operation in Figure~\ref{fig:fig4}.

Finally, the SNS module learns the unstable temporal consistency feature $Z_t^{\mathrm{unst}} \in \mathbb{R}^{C \times \frac{H}{8} \times \frac{W}{8}}$ by concatenating and refining the first-frame depth feature $Z_0^d$ and the aligned feature $Z^{align}$. The same process is repeated in the reverse direction.

\subsection{Masked Stable Module}

\begin{figure}[t]	
\setlength
{\belowcaptionskip}{-5pt}	
\begin{center}		
\includegraphics[width=1.0\linewidth]{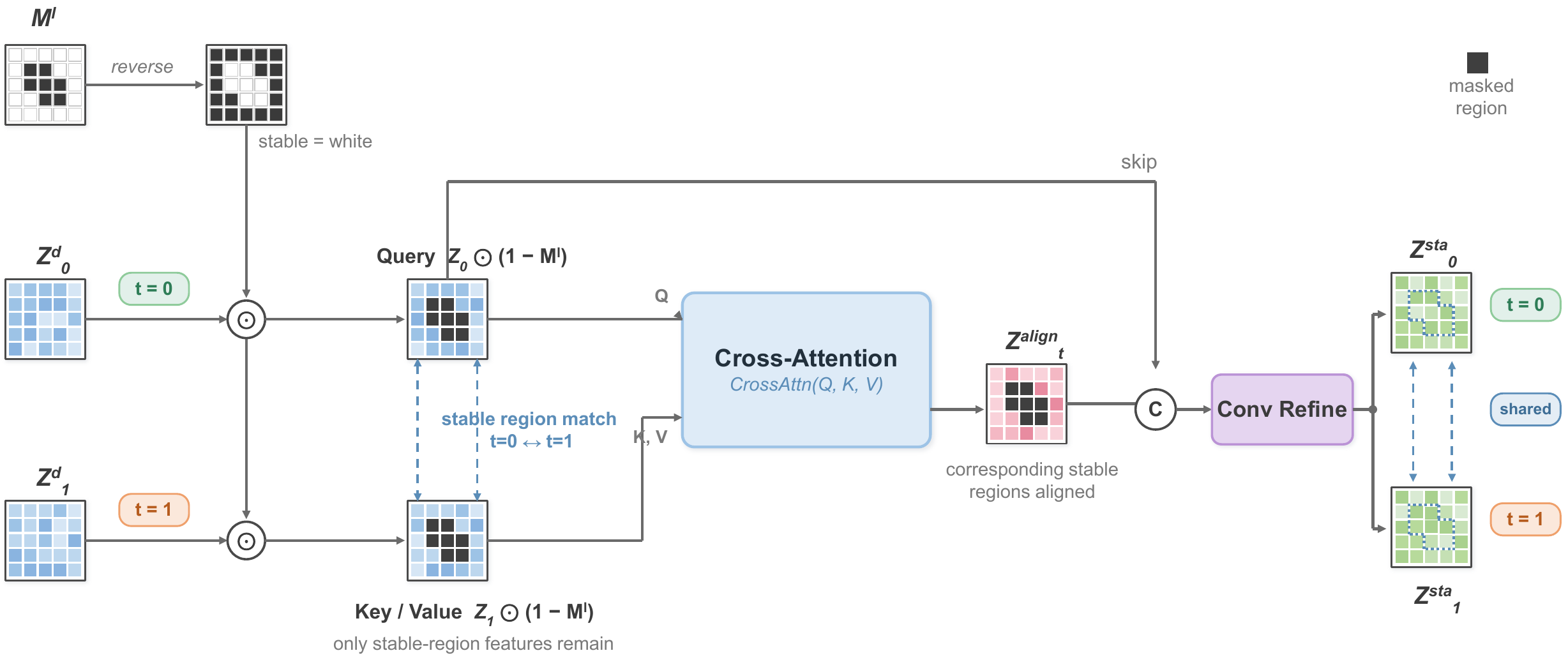}		\caption{{The structure of the MS module. First, an attention mechanism is applied using masked features that retain only geometrically stable regions. Next, a refinement process is conducted to integrate the aligned feature with the depth feature.}}		
\label{fig:fig5}	
\end{center}
\end{figure}

As shown in Figure~\ref{fig:fig5}, the detailed structure of the MS module
is presented. In contrast to the SNS module, the MS module
targets geometrically stable regions denoted by $(1 - M^l)$, where
inter-frame surface-normal variation is small. Because these regions
maintain high inter-frame correlation, explicit correspondence search is
unnecessary, and temporal consistency can be learned directly through
attention between the two frames.
Multiplying $Z_t^d$ with $(1 - M^l)$ retains only the geometrically
stable-region features in each frame. Therefore, we first apply the mask
($1 - M^l$) to the depth context feature $Z_t^d$ of each
frame, which restricts temporal learning to regions where
geometry is preserved across frames.
Then, the MS module follows a cross-attention structure
$CrossAttn(Q,K,V)$. To capture the correlation between the frames, the query feature is derived from $Z_0^d$, while the key and value features are taken from $Z_1^d$. 
After applying feature masking, we additionally use a key-padding mask during cross-attention to exclude geometrically unstable key positions from the softmax computation, ensuring that only stable key positions contribute to the attention output.
This process yields an aligned representation of
corresponding geometrically stable regions across the two
frames, and we repeat this procedure in reverse order for the frames in the
same manner to obtain aligned stable features.
Finally, the aligned features and query features are concatenated and
processed through a refinement process using a simple convolutional
structure. Therefore, we obtain the stable temporal
consistency feature
$Z_t^{\mathrm{sta}} \in \mathbb{R}^{C \times \frac{H}{8} \times \frac{W}{8}}$.
This process is expressed as follows:
\begin{equation} 
\begin{aligned}
    Z_t^{\mathrm{sta}} = conv(concat(Z_t^{align}, Z_t^d\odot (1-M^l))),
\end{aligned}
\end{equation}
where $Z_t^{align}$ denotes the aligned stable feature.

The two modules are complementary by design. Since $M^l$ and
$(1-M^l)$ partition each frame, each module specializes in one region type:
the MS module preserves continuity over the dominant stable areas such as
roads and walls, whereas the SNS module recovers correspondences in the
relatively small but challenging unstable areas such as object boundaries
and moving structures, so neither module alone covers the entire scene.
Then, the independently learned features, $Z_t^{\mathrm{sta}}$
and $Z_t^{\mathrm{unst}}$, are combined to form a unified video feature
$Z_t^v$, enabling consistent prediction over the full frame.

\subsection{Loss function}
Following previous works~\cite{yuan2022new, lee2019big}, we use a scaled Scale-Invariant loss (SILog)~\cite{eigen2014depth} for depth estimation. We use Mean Squared Error (MSE) to supervise both the surface normal and the refined difference mask $M^l$. The mask loss is defined as
\begin{equation}
\begin{aligned}
    L_{mask} = \frac{1}{N} \sum_{i=1}^{N} (M^d(i) - M^l(i))^2,
\end{aligned}
\end{equation}
where $N$ is the number of pixels in the difference mask. The total loss is then given by
\begin{equation}
\begin{aligned}
    L_{total} = L_{Depth} 
    + \lambda_N \cdot L_{Normal} 
    + \lambda_M \cdot L_{Mask},
\end{aligned}
\end{equation}
where $L_{Depth}$ is SILog, and $L_{Normal}$ and $L_{Mask}$ are MSE losses. Following~\cite{lee2019big}, we set the SILog parameter to $\lambda=0.85$ and use $\lambda_N=10$ and $\lambda_M=10$ in all experiments.

\subsection{Evaluation Metrics}

We evaluate our method using both depth accuracy and temporal consistency metrics. 
For depth accuracy, we report standard metrics, including Average Relative Error (Abs Rel), Square Relative Error (Sq Rel), Root Mean Squared Error (RMSE), and Threshold Accuracy ($\delta$) under thresholds $1.25$, $1.25^2$, and $1.25^3$, following prior works.

To evaluate temporal consistency, we adopt the optical flow-based warping metric proposed by Li et al.~\cite{li2021enforcing}, which has also been used in prior video depth estimation methods such as FMNet~\cite{wang2022less}. The corresponding aTC and rTC results are reported in Table~\ref{tab:tab4}. The metrics are defined as
\begin{equation*}
\begin{aligned}
    aTC_t &= \frac{1}{\sum (K_t == 1)} \sum K_t \left| \frac{D_t - D_t^w}{D_t} \right|, \\
    rTC_t &= \frac{1}{\sum (K_t == 1)} \sum K_t \left[ \text{Max} \left( \frac{D_t}{D_t^w}, \frac{D_t^w}{D_t} \right) < \text{thr} \right],
\end{aligned}
\end{equation*}
where $D_t$ is the predicted depth, $D_t^w$ is the warped depth from $D_{t-1}$, and $K_t$ is a depth validity mask. 

In addition, we also report the optical flow-based warping error (OPW), following prior works such as FMNet~\cite{wang2022less}. Specifically, OPW measures the discrepancy between the predicted depth and the warped depth:
\begin{equation*}
    \text{OPW}_t = \frac{1}{\sum (K_t == 1)} \sum K_t \left| D_t - D_t^w \right|.
\end{equation*}

For depth warping, we use FlowFormer~\cite{huang2022flowformer} to estimate optical flow. The OPW results are reported in Table~\ref{tab:tab3}.

\section{Experiments}

\subsection{Implementation Details}

The proposed method is implemented using the AdamW~\cite{loshchilov2017decoupled} optimizer with $\beta$ parameters of 0.9 and 0.999, a batch size of 2 with 2 frames per batch, and a weight decay set to $10^{-2}$. All attention modules are implemented with 8 heads. For frame-pair sampling, we use two temporally adjacent frames as the input pair during training and standard evaluation. During inference, STATIC is applied to consecutive frame pairs using the same two-frame formulation, and the predicted depth maps are evaluated following the standard benchmark protocol. We train the model for 20 epochs on each dataset, starting from an initial learning rate of $4\times10^{-5}$, which linearly decays to $4\times10^{-6}$ over the training process. With four NVIDIA 4090 GPUs, each epoch takes approximately 60 minutes. The encoder backbone is initialized with pre-trained SwinV2-L~\cite{liu2022swin} weights. During testing, final depth predictions are obtained by averaging the outputs from both the original and mirrored inputs. We follow the training and testing protocols used in~\cite{agarwal2023attention, lee2019big}.

\subsection{Datasets}

\textbf{KITTI Eigen}: The KITTI dataset~\cite{geiger2012we} is a widely used benchmark for outdoor depth estimation. We adopt the Eigen split, which includes 23,488 training images and 697 test images. The corresponding video sequences are used, with each frame having a resolution of 375$\times$1241 pixels. For evaluation, we apply the Garg crop~\cite{garg2016unsupervised} and limit depth to 80 meters.

\setlength{\parindent}{0pt}\textbf{NYU Depth v2}: NYU v2~\cite{silberman2012indoor} is an indoor dataset containing 120,000 RGB-depth pairs with 480$\times$640 resolution across 464 indoor scenes. For video-based evaluation, we follow the protocol of~\cite{wang2023neural}. Specifically, we use 249 scenes with RGB and synchronized depth pairs~\cite{lee2019big} for training, and the remaining 215 scenes with 654 images for testing. Depth is capped at 10 meters, and center cropping is applied following Eigen et al.~\cite{eigen2014depth}.

\setlength{\parindent}{0pt}\textbf{MPI-Sintel}: We evaluate our method on MPI-Sintel~\cite{butler2012naturalistic}, which contains 23 synthetic dynamic-scene sequences with metric-scale ground-truth depth. The dataset provides clean and final rendering variants, with the final version including realistic effects such as motion blur and haze.

\begin{table*}[ht]
\centering
\setlength{\tabcolsep}{1pt}
\resizebox{1.0\linewidth}{!}{
\begin{tabular}{lccccccccccc}
\toprule
\textbf{Method} & \textbf{Frame} & \textbf{C} & \textbf{O} & \textbf{M}& \textbf{Abs Rel $\downarrow$} & \textbf{Sq Rel $\downarrow$} & \textbf{RMSE $\downarrow$} & \textbf{ $\delta < 1.25$  $\uparrow$} & \textbf{ $\delta < 1.25^2$  $\uparrow$} & \textbf{ $\delta < 1.25^3$ $\uparrow$} \\
\midrule
    PEM ~\cite{lee2022edgeconv}      & SF & - & - & -    & 0.068 & 0.221 & 2.127 & 0.958 & 0.993 & 0.9983 \\
    CATNet ~\cite{tang2024catnet}  & SF & - & - & -    & 0.052 & 0.160 & 2.153 & 0.972 & 0.997 & 0.9990 \\
    SSRDepth ~\cite{zou2025ssrdepth}  & SF & - & - & -    & 0.053 & 0.157 & 2.125 & 0.973 & 0.997 & 0.9990 \\
    AdaBins ~\cite{bhat2021adabins}        & SF & - & - & -    & 0.058 & 0.190 & 2.360 & 0.964 & 0.995 & 0.9993 \\
    BinsFormer ~\cite{li2024binsformer}       & SF & - & - & -         & 0.058 & 0.190 & 2.336 & 0.964 & 0.996 & 0.9994 \\
    DepthFormer ~\cite{agarwal2022depthformer}     & SF & - & - & -           & 0.053 & 0.187 & 2.285 & 0.970 & 0.996 & 0.9994 \\
    PixelFormer ~\cite{agarwal2023attention}     & SF & - & - & -         & 0.052 & 0.152 & 2.093 & 0.975 & 0.997 & 0.9994 \\
    GEDepth ~\cite{yang2023gedepth} & SF & $\checkmark$ & - & - & 0.049 & 0.143 & 2.048 & 0.976 & 0.997 & 0.9994 \\
    \midrule
    NeuralRGB ~\cite{liu2019neural}      & MF & $\checkmark$  & - & -           & 0.100 & -     & 2.829 & -     & -     & -     \\
    ST-CLSTM ~\cite{zhang2019exploiting}       & MF & -  & - & $\checkmark$         & 0.101 & -     & 4.137     & 0.890 & 0.970 & 0.9890 \\
    FlowGRU ~\cite{eom2019temporally}       & MF & -  & $\checkmark$ & $\checkmark$      & 0.112 & 0.700 & 4.260 & 0.936 & 0.983 & 0.9930 \\
    Flow2Depth ~\cite{xie2020video}    & MF & $\checkmark$ & $\checkmark$ & -      & 0.109 & -     & 4.284 & 0.910 & 0.980 & 0.9900 \\
    RDE-MV ~\cite{patil2020don}      & MF & - & - & $\checkmark$        & 0.111 & 0.821 & 4.650 & 0.898 & 0.972 & 0.9890 \\
    FMNet ~\cite{wang2022less}        & MF & - & - & -       & 0.069 & 0.342 & 3.340 & 0.946 & 0.986 & 0.9960 \\
    ManyDepth-FS ~\cite{watson2021temporal}      & MF & $\checkmark$ & - & -         & 0.053 & 0.243 & 2.248 & 0.975 & 0.997 & 0.9994 \\
    TC-Depth-FS ~\cite{ruhkamp2021attention}      & MF & - & - & -          & 0.059 & 0.249 & 3.280 & 0.947 & 0.985 & 0.9940 \\
    MAMo  ~\cite{yasarla2023mamo}    & MF & - & $\checkmark$ & $\checkmark$     & 0.049 & 0.130 & 1.989 & 0.977 & 0.998 & 0.9995 \\
    FutureDepth  ~\cite{yasarla2024futuredepth}    & MF & - & - & -     & 0.044 & 0.119 & 1.920 & 0.983 & \textbf{0.999} & \textbf{0.9997} \\
    \midrule
    \textbf{STATIC}    & MF & - & - & -  & \textbf{0.041} & \textbf{0.117} & \textbf{1.917} & \textbf{0.988} & \textbf{0.999} & \textbf{0.9997} \\
\bottomrule

\end{tabular}
}
\caption{Performance comparison between various methods on the KITTI Eigen dataset. The best results are in bold. "MF" indicates multi-frame methods, "SF" indicates single-frame methods, "O" represents optical flow, "C" represents camera parameter, and "M" represents memory. Baseline results are cited from the corresponding published papers.}
\label{tab:tab1}
\end{table*}

\subsection{Results}

\textbf{Results on KITTI Eigen}: Table~\ref{tab:tab1} presents our results on the KITTI dataset, where STATIC demonstrates robust performance with minimal input requirements. We compare our approach against various state-of-the-art multi-frame and single-frame methods. Even without relying on explicit camera parameters, heavy multi-frame memory banks, or external optical flow estimators during inference, our model achieves strong performance across all metrics. Our model's depth embedder is based on PixelFormer~\cite{agarwal2023attention} and substantially outperforms the baseline, highlighting the effectiveness of the temporal consistency module. Furthermore, our approach improves RMSE by approximately 3.62\% over the recently proposed MAMo~\cite{yasarla2023mamo}, even though MAMo utilizes more frames. Our method also surpasses FutureDepth~\cite{yasarla2024futuredepth} in the major error metrics, achieving a 6.82\% reduction in Abs Rel, a 1.68\% reduction in Sq Rel, and a 0.16\% reduction in RMSE, while improving $\delta < 1.25$ from 0.983 to 0.988 and matching the best results on $\delta < 1.25^2$ and $\delta < 1.25^3$. These results indicate that separately modeling geometrically stable and unstable regions helps maintain consistent depth predictions in outdoor scenes with noticeable inter-frame geometric variation.

\begin{table}[t]
\setlength{\belowcaptionskip}{-5pt}
\centering
\setlength{\tabcolsep}{12pt}
\resizebox{0.8\linewidth}{!}{
\begin{tabular}{cccc}
\hline
\textbf{Type} & \textbf{Method}           & \textbf{$\delta < 1.25$  $\uparrow$} & \textbf{Abs Rel $\downarrow$} \\ \hline
SF   
    & SSRDepth~\cite{zou2025ssrdepth} & 0.933 & 0.086   \\
    & MiDas-v2.1-Large~\cite{ranftl2020towards} & 0.910 & 0.095   \\
     & DPT-Large  ~\cite{ranftl2021vision}       & 0.928 & 0.084   \\ \hline
      & WSVD  ~\cite{wang2019web}           & 0.768 & 0.164   \\
     & ST-CLSTM ~\cite{zhang2019exploiting}         & 0.833 & 0.131   \\
MF    & DeepV2D   ~\cite{teed2018deepv2d}       & 0.924 & 0.082 \\
          & FMNet ~\cite{wang2022less}            & 0.832 & 0.134   \\
     & VITA  ~\cite{xian2023vita}           & 0.922 & 0.092   \\
     & MAMo ~\cite{yasarla2023mamo}          & 0.942 & 0.074    \\ 
     & NVDS ~\cite{wang2023neural} &0.950 &0.072 \\ 
     & Video Depth Anything-S$^{\dagger}$~\cite{chen2025video} & 0.959 & 0.077 \\ 
     \hline
MF   & \textbf{STATIC}          &\textbf{0.954} & \textbf{0.071} \\
\hline
\end{tabular}
}
\caption{Performance comparison between various methods on the NYU v2 dataset. Baseline results are cited from the corresponding published papers. $^{\dagger}$: trained with large-scale pretraining under a different supervision setting.}
\label{tab:tab2}
\end{table}

\setlength{\parindent}{0pt}\textbf{Results on NYU v2}: Table~\ref{tab:tab2} presents our results on the NYU v2 dataset. Following the evaluation procedure of~\cite{agarwal2023attention}, our method achieves the best performance on both $\delta < 1.25$ and Abs Rel among the compared methods.
Specifically, STATIC improves $\delta < 1.25$ from 0.950 to 0.954 over NVDS~\cite{wang2023neural}, corresponding to a 0.42\% gain, and from 0.942 to 0.954 over MAMo, corresponding to a 1.27\% gain. In terms of Abs Rel, our method also outperforms NVDS by reducing the error from 0.072 to 0.071, while further improving over MAMo from 0.074 to 0.071. These results show that STATIC provides consistent improvements in both threshold accuracy and relative error over the compared recent multi-frame methods on indoor video depth estimation.

\setlength{\parindent}{0pt}\textbf{Results on Sintel}: Table~\ref{tab:tab3} presents our results on the Sintel dataset~\cite{butler2012naturalistic}. Following the evaluation procedure of~\cite{wang2023neural}, we compare against representative image, flow, camera, and memory-based baselines. Our method achieves the best performance among the compared methods in $\delta < 1.25$, $Rel$, and $OPW$, indicating strong depth accuracy and temporal consistency on Sintel. Compared with prior video-based approaches such as NVDS~\cite{wang2023neural}, STATIC obtains a lower $OPW$, suggesting improved temporal stability. These results indicate that the proposed surface-normal-guided temporal modeling improves temporal consistency while maintaining competitive depth accuracy.

\subsection{Selection of Comparable Baselines}

In this section, we discuss recent methods excluded from our main comparison due to differences in training settings.

Depth Anything V2~\cite{yang2024depth} uses web-scale pretraining, while DepthCrafter~\cite{hu2025depthcrafter} and ECoDepth~\cite{patni2024ecodepth} leverage diffusion-based frameworks and large-scale pretraining, making direct comparisons difficult.

STATIC also uses Depth Anything V2 to generate auxiliary surface-normal targets, as sparse depth GT, particularly in KITTI~\cite{geiger2012we}, cannot provide reliable dense normals for adjacent frame pairs through Sobel filtering. However, the primary depth objective is supervised by ground-truth depth during training, and STATIC is evaluated separately on each benchmark dataset without directly using web-scale data for training. Although pretrained knowledge may indirectly contribute through the auxiliary targets, the supervision settings are not strictly equivalent.

We also consider geometry-based methods such as LightedDepth~\cite{zhu2023lighteddepth}, which rely on optical flow or pose estimation and may propagate estimation errors. As shown in Table~\ref{tab:tab3}, STATIC achieves better accuracy and temporal consistency than Robust-CVD and GEDepth without external motion estimation at inference.

\begin{table}[]
\setlength{\belowcaptionskip}{-5pt}
\setlength{\tabcolsep}{3pt}
\centering
\setlength{\tabcolsep}{10pt}
\resizebox{0.85\linewidth}{!}{
\small
\renewcommand{\arraystretch}{0.9}
\centering
\begin{tabular}{c|c|cccc}
\toprule
\textbf{Method}     & \textbf{Type} & \textbf{$\delta_1\uparrow$} & \textbf{$Rel\downarrow$ }  & \textbf{$OPW\downarrow$} & \textbf{$\textit{T}(ms)$} \\ 
\midrule
\midrule
DPT ~\cite{ranftl2021vision}      & \multirow{2}{*}{Image}  & \underline{0.597} & 0.339 & 0.612 & 67.7 \\
MiDas ~\cite{ranftl2020towards}   &                         & 0.485 & 0.410 & 0.843 & 53.0 \\
\midrule
CVD   ~\cite{luo2020consistent}    & \multirow{2}{*}{Flow}   & 0.518 & 0.406 & 0.497 & -\\
WSVD  ~\cite{wang2019web}   &                         & 0.501 & 0.439 & 0.577 & 296.6\\
\midrule
Robust-CVD ~\cite{kopf2021robust} & \multirow{2}{*}{Camera} & 0.521 & 0.422 & 0.475 & - \\
GEDepth$^\dagger$  ~\cite{yang2023gedepth} &                         & 0.522 & 0.430 & 0.473 &185\\
\midrule
ST-CLSTM ~\cite{zhang2019exploiting}  & Memory                  & 0.351 & 0.517 & 0.585 &40.5\\
\midrule
FMNet ~\cite{wang2022less}    & \multirow{4}{*}{Video}  & 0.357 & 0.513 & 0.521 &270.3 \\
DeepV2D ~\cite{teed2018deepv2d}   &                         & 0.486 & 0.526 & 0.534 &975.68 \\
NVDS-MiDas ~\cite{wang2023neural} &                         & 0.532 & 0.374 & 0.469 &108.4 \\
NVDS-DPT  &                         & 0.591 & \underline{0.335} & \underline{0.424} &120.76 \\
\midrule
\midrule
STATIC       & Video                   & \textbf{0.598} & \textbf{0.330} & \textbf{0.405} & 197.32 \\
\bottomrule
\end{tabular}
}
\caption{Performance comparison of various methods on the Sintel dataset. The best results are highlighted in \textbf{bold}, and the second best results are \underline{underlined}. Note that $^\dagger$ indicates retrained model and \textit{T} denotes inference time. Baseline results are cited from the corresponding published papers.}
\label{tab:tab3}
\end{table}

\begin{figure*}[t]
	\setlength{\belowcaptionskip}{-5pt}
	\begin{center}
		\includegraphics[width=1.0\linewidth]{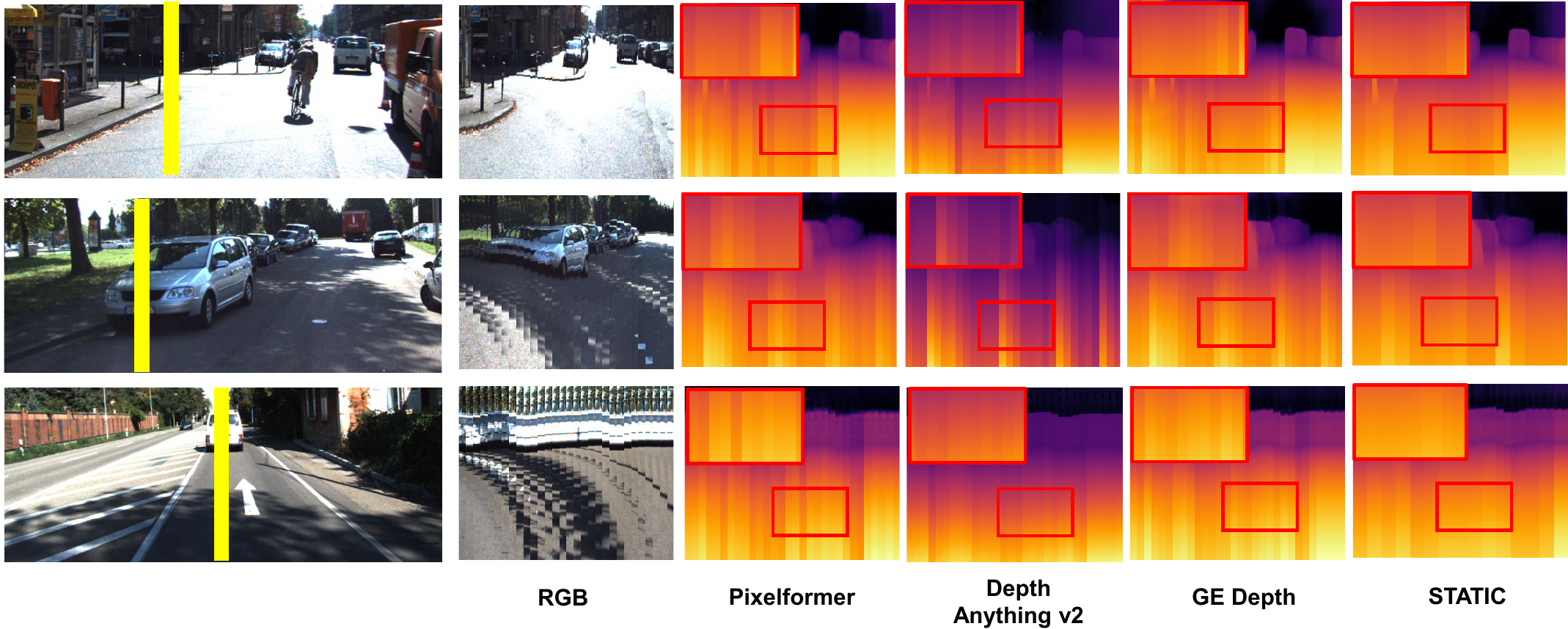}
		\caption{Temporal consistency comparison on the KITTI Eigen dataset. 
The temporal profiles are generated from the marked slice regions in the RGB frames to visualize frame-to-frame depth continuity. 
The red boxes indicate zoomed regions where temporal discontinuities are more clearly observed.}
		\label{fig:fig6}
	\end{center}
\end{figure*}

\begin{table}[t]
\setlength{\belowcaptionskip}{-5pt}
\centering
\setlength{\tabcolsep}{12pt}
\resizebox{0.7\linewidth}{!}{
\begin{tabular}{cccc}
\hline
\textbf{Type} & \textbf{Method}        & \textbf{rTC} $\uparrow$   & \textbf{aTC} $\downarrow$   \\ \hline
     & NeWCRFs ~\cite{yuan2022new}      & 0.914 & 0.116 \\
SF   & iDisc~\cite{piccinelli2023idisc}         & 0.923 & 0.108 \\
     & GEDepth ~\cite{yang2023gedepth}     & 0.919 & 0.133 \\ 
     & Depth Anything V2 ~\cite{yang2024depth} & 0.946 & 0.099 \\ \hline
     & TC-Depth-FS ~\cite{ruhkamp2021attention}   & 0.901 & 0.122 \\
MF   & Many-Depth-FS ~\cite{watson2021temporal} & 0.920 & 0.111 \\
     & NVDS  ~\cite{wang2023neural}     & 0.951 & 0.096 \\
     & MAMo ~\cite{yasarla2023mamo}       & 0.966 & 0.086 \\ \hline
MF   & \textbf{STATIC}        & \textbf{0.983} & \textbf{0.075} \\ \hline
\end{tabular}
}
\caption{Comparison of temporal consistency across various methods on rTC and aTC.}
\label{tab:tab4}
\end{table}

\subsection{Temporal Consistency on KITTI Eigen}
Table~\ref{tab:tab4} presents a numerical comparison across various methods using standard evaluation metrics to assess temporal consistency. Notably, our approach achieves a 1.76\% improvement in relative Temporal Consistency (rTC) compared to the previous highest performance. Moreover, a 12.79\% relative improvement is observed in absolute Temporal Consistency (aTC). These findings suggest that independently modeling geometrically stable and unstable regions contributes to improved temporal consistency, even when only two adjacent frames are used.
Further insights are provided by Figure~\ref{fig:fig6}, which shows qualitative comparisons across 60 consecutive frames, illustrating each method's ability to maintain temporal consistency over time. Competing methods often display striping artifacts, indicating instability in estimated depth. In contrast, STATIC shows a reduction in both the occurrence and intensity of these artifacts, highlighting its stability in maintaining temporal consistency over an extended sequence. These qualitative results suggest that STATIC produces more stable depth predictions, leading to smoother frame-to-frame transitions and improved visual coherence.

\begin{figure*}[t]
	\setlength{\belowcaptionskip}{-5pt}
	\begin{center}
		\includegraphics[width=1.0\linewidth]{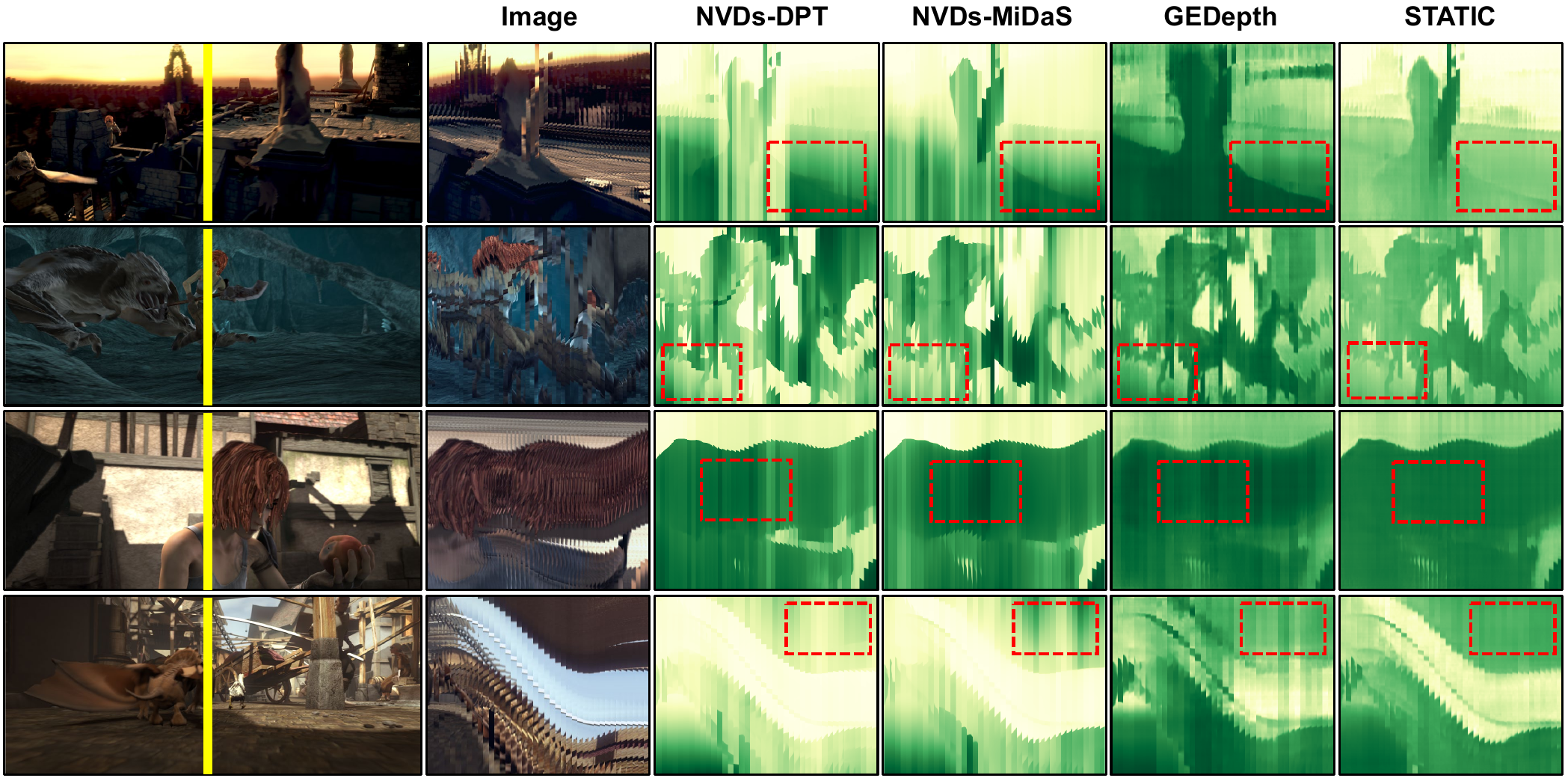}
        \captionof{figure}{Qualitative comparison of video methods on the MPI-Sintel dataset. The temporal profiles are generated from the marked slice regions in the RGB frames to visualize frame-to-frame depth continuity. The red boxes highlight regions where differences in temporal stability are noticeable.}
    	\label{fig:fig7}
    \end{center}
\end{figure*}

\subsection{Temporal Consistency on MPI-Sintel}
Table~\ref{tab:tab3} presents quantitative results on the Sintel dataset. In addition to depth accuracy metrics, we report the optical flow-based warping error (OPW) defined in Section 3.7, where lower values indicate better temporal stability. Among the compared methods, our approach achieves the best values in $\delta$, Rel, and OPW. In particular, the lower OPW suggests improved temporal consistency across video frames.

Figure~\ref{fig:fig7} provides a qualitative comparison across all frames of each scene to illustrate temporal consistency. While the compared approaches often exhibit striping patterns that reflect temporal inconsistencies in the estimated video depth, STATIC reduces both the frequency and intensity of these artifacts. These results suggest that our model produces more stable depth predictions, leading to smoother frame-to-frame transitions and improved visual coherence.

\begin{figure*}[t]
	\setlength{\belowcaptionskip}{-10pt}
	\begin{center}
		\includegraphics[width=1.0\linewidth]{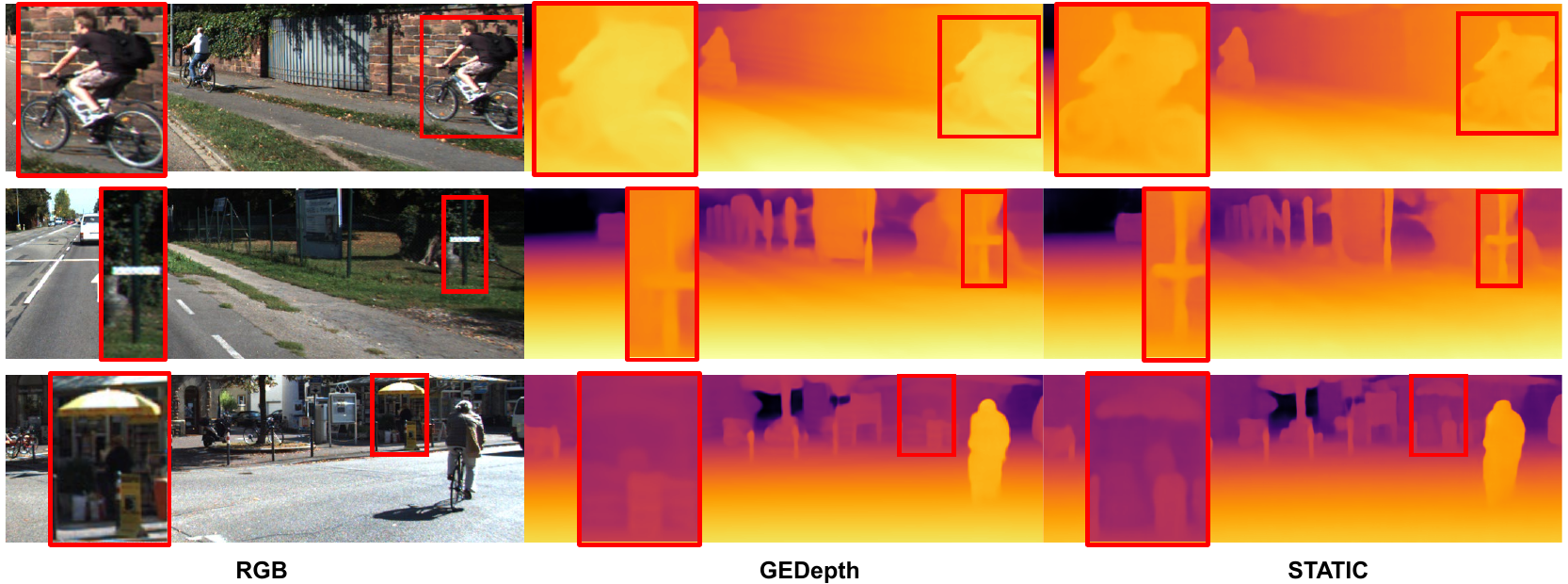}
		\caption{ Qualitative comparison of video methods on the KITTI Eigen dataset. }
		\label{fig:fig8}
	\end{center}
\end{figure*}

\begin{figure*}[t]
	\setlength{\belowcaptionskip}{-10pt}
	\begin{center}
		\includegraphics[width=0.8\linewidth]{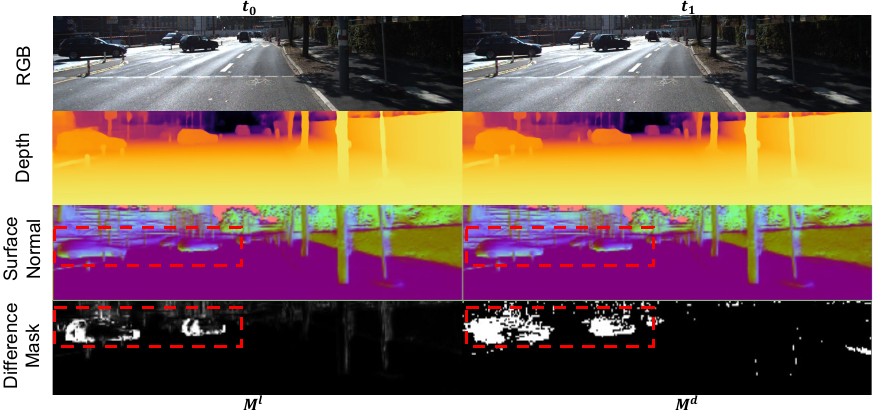}
        \captionof{figure}{Results of STATIC in a scenario where the camera is stationary, and only the object is moving.}
        \label{fig:sup_fig1}
    \end{center}
    \setlength{\belowcaptionskip}{-15pt}
    \begin{center}
		\includegraphics[width=0.8\linewidth]{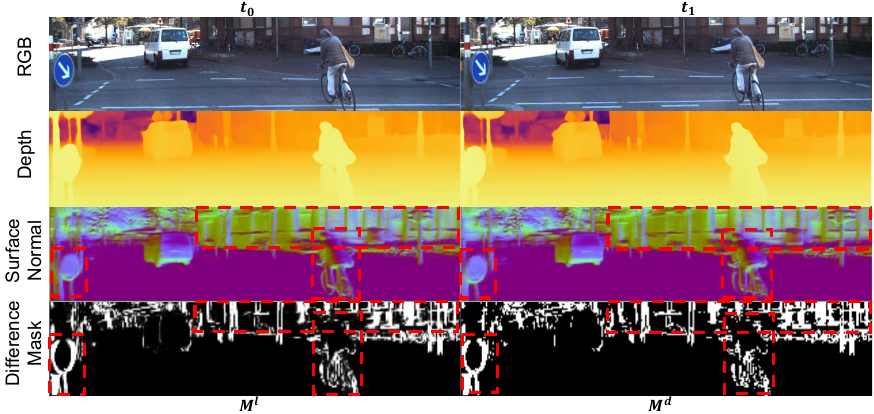}
        \captionof{figure}{Results of STATIC in a scenario where both the camera and the object are moving.}
        \label{fig:sup_fig2}
    \end{center}
\end{figure*}

\subsection{Qualitative Results}

Figure~\ref{fig:fig8} presents qualitative comparisons with other video-based methods. The proposed mask $M^l$ highlights regions with noticeable geometric changes, including small and distant objects such as the parasol. By separating geometrically stable and unstable regions, STATIC better preserves object structures, such as the bicycle, person, and sign, particularly around object boundaries. Similar improvements are observed on MPI-Sintel in Figure~\ref{fig:fig7}, where STATIC produces clearer object boundaries and more coherent depth estimates, even in scenes with complex geometric variations.

Figures~\ref{fig:sup_fig1} and~\ref{fig:sup_fig2} further illustrate the behavior of the proposed mask under different motion scenarios. With a stationary camera, the mask primarily highlights moving objects while suppressing geometrically stable background regions. When both the camera and objects move, the mask also captures geometric variation in the background and foreground. These observations indicate that the proposed mask responds to inter-frame geometric changes induced by both object and camera motion, enabling region-specific temporal processing while preserving structural consistency.

\section{Ablation Study}

\subsection{Effect of MS module} 
As shown in Table~\ref{tab:tab5}, the MS module mainly improves temporal consistency in geometrically stable regions. Since geometrically stable regions occupy most of the scene, removing the MS module increases the areas where temporal consistency cannot be properly learned, resulting in a 1.77\% increase in RMSE. This result highlights the importance of the MS module in maintaining overall performance.

\subsection{Effect of SNS module}
Table~\ref{tab:tab5} also shows the contribution of the SNS module, which improves temporal consistency in geometrically unstable region by modeling correspondences in contours and small regions with significant inter-frame geometric variation. Because these geometrically unstable regions occupy a relatively small portion of the image, removing the SNS module causes a smaller RMSE increase of 1.15\% compared to removing the MS module. Since the model learns geometrically stable and unstable regions independently and then integrates them, the absence of either module leaves certain regions insufficiently modeled, reducing overall performance. These ablation results confirm that both modules are complementary and that their joint use is essential for achieving the best performance.

\begin{table}[t]
\setlength{\belowcaptionskip}{-5pt}
\centering
\setlength{\tabcolsep}{10pt}
\resizebox{0.72\linewidth}{!}{
\begin{tabular}{ccccc}
\hline
\textbf{SNS} & \textbf{MS}  & \textbf{Abs Rel $\downarrow$} & \textbf{RMSE $\downarrow$} & \textbf{$\delta < 1.25$  $\uparrow$} \\ \hline
-          & -         &   0.051    &   2.022    &  0.977     \\
$\checkmark$          & -         &   0.047    &   1.951       &    0.982  \\ 
-          & $\checkmark$         &   0.045    &   1.939       &    0.984  \\ 
$\checkmark$          & $\checkmark$     &  \textbf{0.041}  &   \textbf{1.917}  & \textbf{0.988}   \\ \hline
\end{tabular}
}
\caption{The ablation experiment evaluates the individual effects of the SNS module and the MS module on the KITTI Eigen dataset.}
\label{tab:tab5}
\end{table}

\begin{figure*}[t]
	\setlength{\belowcaptionskip}{-5pt}
	\begin{center}
		\includegraphics[width=1.0\linewidth]{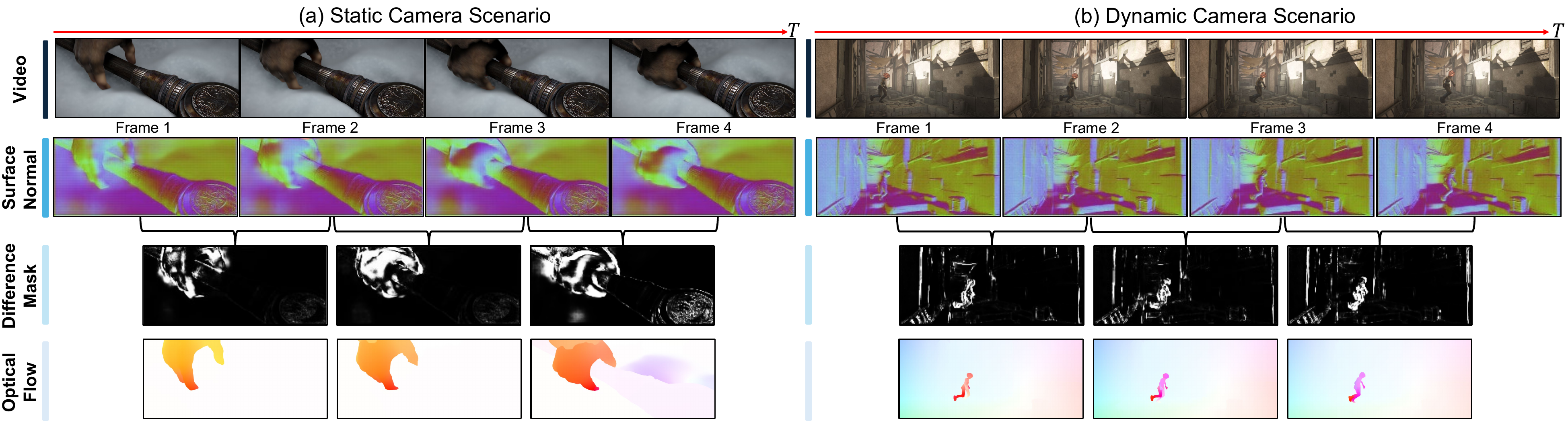}
		\caption{Qualitative comparison under stationary-camera and moving-camera scenarios. Video frames, surface normal, and the corresponding difference mask across consecutive frames are shown, with optical flow for comparison.}
		\label{fig:fig11}
	\end{center}
\end{figure*}

\begin{table}[t]
\setlength{\belowcaptionskip}{-5pt}
\centering
\setlength{\tabcolsep}{10pt}
\resizebox{0.6\linewidth}{!}{
\begin{tabular}{c|ccc}
\toprule
\textbf{Frame Speed} & \textbf{$\delta_1\uparrow$} & \textbf{$Rel\downarrow$}   & \textbf{$OPW\downarrow$} \\ \midrule\midrule
$\times1$       &     0.598   & 0.330    & 0.405    \\
$\times 0.5$      &    0.595  & 0.334   &  0.411  \\
$\times 0.33$      &   0.571   & 0.359  & 0.424   \\ \bottomrule
\end{tabular}
}
\caption{Performance Evaluation at Different Frame Speeds.}
\label{tab:tab6}
\end{table}

\subsection{Robustness of the Difference Mask}

Figure~\ref{fig:fig11}(a) further compares the proposed difference mask with optical flow under stationary-camera and moving-camera conditions. In Figure~\ref{fig:fig11}(a), the mask highlights the hand region with noticeable inter-frame geometric variation while leaving the stable background largely unchanged, and in Figure~\ref{fig:fig11}(b), camera motion introduces additional responses in the background, both showing good agreement with optical flow. 
The key difference lies in what the two cues capture. Optical flow models global scene motion and thus produces broad responses over the entire background, whereas the difference mask responds only to structural changes, as also observed around the cane between frames 3 and 4 in Figure~\ref{fig:fig11}(a). This selectivity allows STATIC to enforce consistency according to the degree of geometric variation, preserving structural fidelity in changed regions while avoiding unnecessary modification of unchanged areas.

\begin{figure}[t]
	\setlength{\belowcaptionskip}{-10pt}
	\begin{center}
		\includegraphics[width=0.78\linewidth]{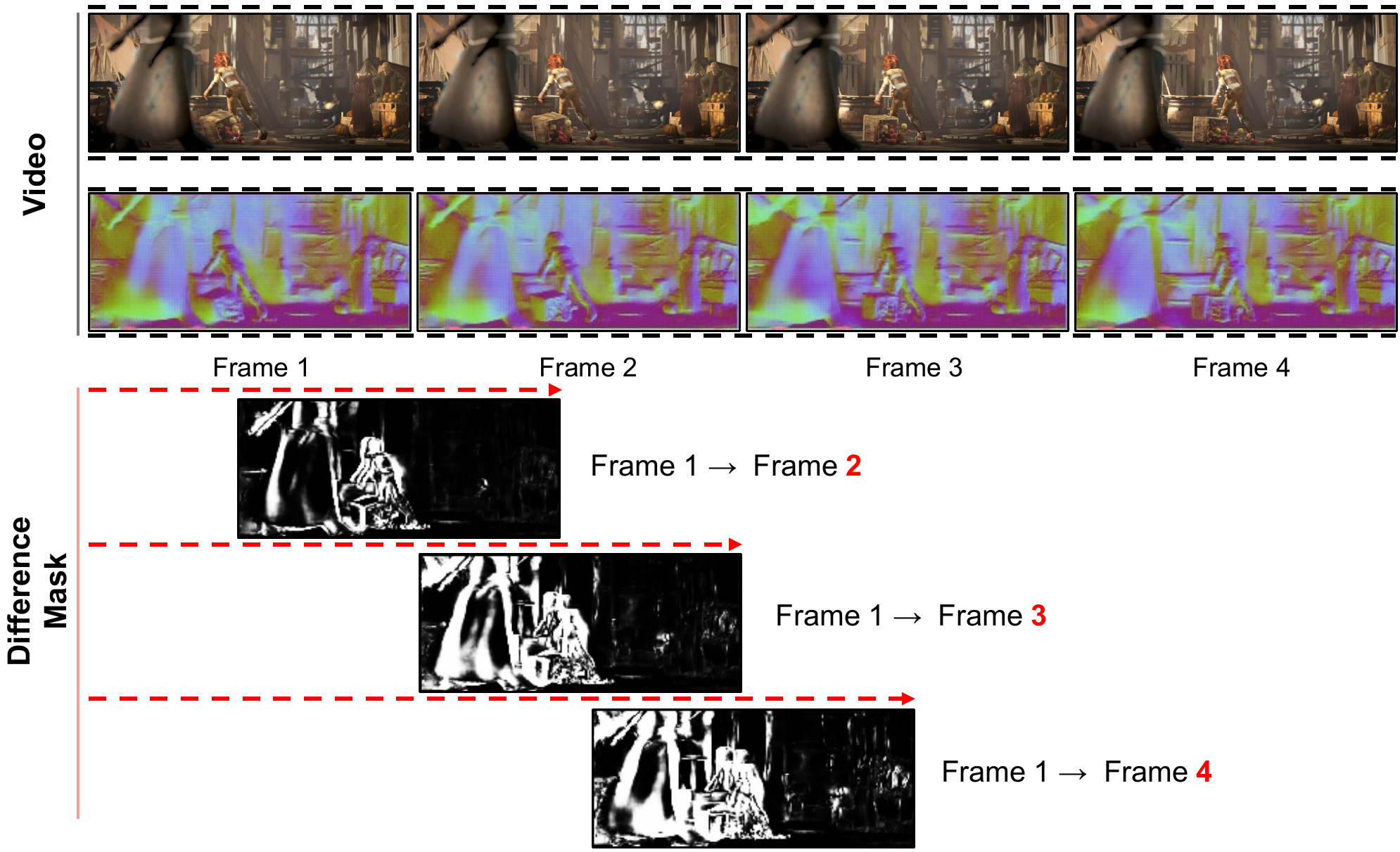}
		\caption{Comparison of Difference Mask Variations with Increased Frame Intervals.}
		\label{fig:fig12}
	\end{center}
\end{figure}

\subsection{Dependency of Frame Rate}

Table~\ref{tab:tab6} evaluates STATIC at different frame speeds by skipping intermediate frames. As the frame interval increases, the overall performance gradually decreases due to larger inter-frame changes. Nevertheless, even at $\times0.33$ frame speed, STATIC maintains a $\delta_1$ of 0.571 and an OPW of 0.424, indicating reasonable robustness to larger temporal gaps. Furthermore, Figure~\ref{fig:fig12} shows that the difference masks continue to capture regions with noticeable geometric variation despite the increased frame intervals. These results suggest that the proposed surface-normal-based mask remains effective across different frame-rate settings, although larger temporal gaps can reduce prediction accuracy and temporal consistency.

\subsection{Computational Cost Analysis}

To analyze the efficiency of STATIC, we compare its computational cost with existing image- and video-based depth estimation methods. All measurements are conducted on a single NVIDIA RTX 3090 GPU. For inference time, we use a 32-frame video clip and report the average time per frame. GPU memory denotes peak inference memory usage.

As shown in Table~\ref{tab:computational_cost}, ours achieves competitive computational efficiency among video-based depth estimation methods, maintaining relatively low FLOPs and peak GPU memory usage while providing comparable per-frame inference speed. This efficiency mainly comes from the design of STATIC, which does not rely on external motion estimation modules such as optical flow or camera-parameter estimation. Instead, it uses internal depth-derived geometric cues and learns temporal consistency from two adjacent frames, improving temporal consistency without heavy multi-frame processing.

\begin{table}[t]
\setlength{\belowcaptionskip}{-10pt}
\centering
\renewcommand{\arraystretch}{1.12}
\resizebox{\linewidth}{!}{
\begin{tabular}{llcccc}
\toprule
\textbf{Type} & \textbf{Method} & \textbf{Params (M)} & \textbf{FLOPs (G)} & \textbf{GPU Mem. (MB)} & \textbf{Inference Time (ms)}\\
\hline
Image & Depth Anything V2 ~\cite{yang2024depth} & 335.316 & 416.815 & 1483 & 38.767 \\
\hline
\multirow{5}{*}{Video} 
& NVDS-MiDaS ~\cite{wang2023neural} & 198.581 & 238.846 & 1469 & 26.459 \\
& NVDS-DPT  & 437.274 & 425.770 & 2378 & 42.831 \\
& GEDepth ~\cite{yang2023gedepth}  & 277.135 & 318.211 & 1692 & 32.812 \\
& Video Depth Anything ~\cite{chen2025video} & 384.390 & 453.725 & 4333 & 30.707 \\
& STATIC & 280.610 & 243.384 & 1626 & 30.553  \\
\hline
\end{tabular}
}
\caption{
Computational cost comparison between image-based and video-based depth estimation methods at an input resolution of $240 \times 480$.}

\label{tab:computational_cost}
\end{table}

\section{Discussion}

\subsection{Surface Normal Ground Truth Generation}

We employ supervised learning for surface normals, which requires ground truth (GT) data. To simplify GT acquisition, we generate surface-normal GT by applying a Sobel filter to depth maps. However, LiDAR-based depth maps are very sparse, making it difficult to apply the Sobel filter effectively. To address this limitation, we first generate dense depth maps using Depth Anything V2~\cite{yang2024depth}.

Specifically, given a dense depth map $D$, we compute depth gradients using Sobel filters $K_x$ and $K_y$, and construct the normalized surface-normal representation as

\begin{equation}
G_x = K_x * D, \quad G_y = K_y * D,
\quad N(p) = \frac{[G_x(p),\,G_y(p),\,1]^\top}
{\sqrt{G_x(p)^2+G_y(p)^2+1}}.
\end{equation}

This image-space construction does not require camera intrinsics but is affected by depth scale. However, since our method uses normal variations as geometric guidance rather than for metric 3D reconstruction, strict scale invariance is not required, although scale changes may influence the difference mask.

This process is performed offline for all training images, and the precomputed surface-normal GT is used for auxiliary supervision during training.

\begin{figure}[t]
	\setlength{\belowcaptionskip}{-15pt}
	\begin{center}
\includegraphics[width=0.8\linewidth]{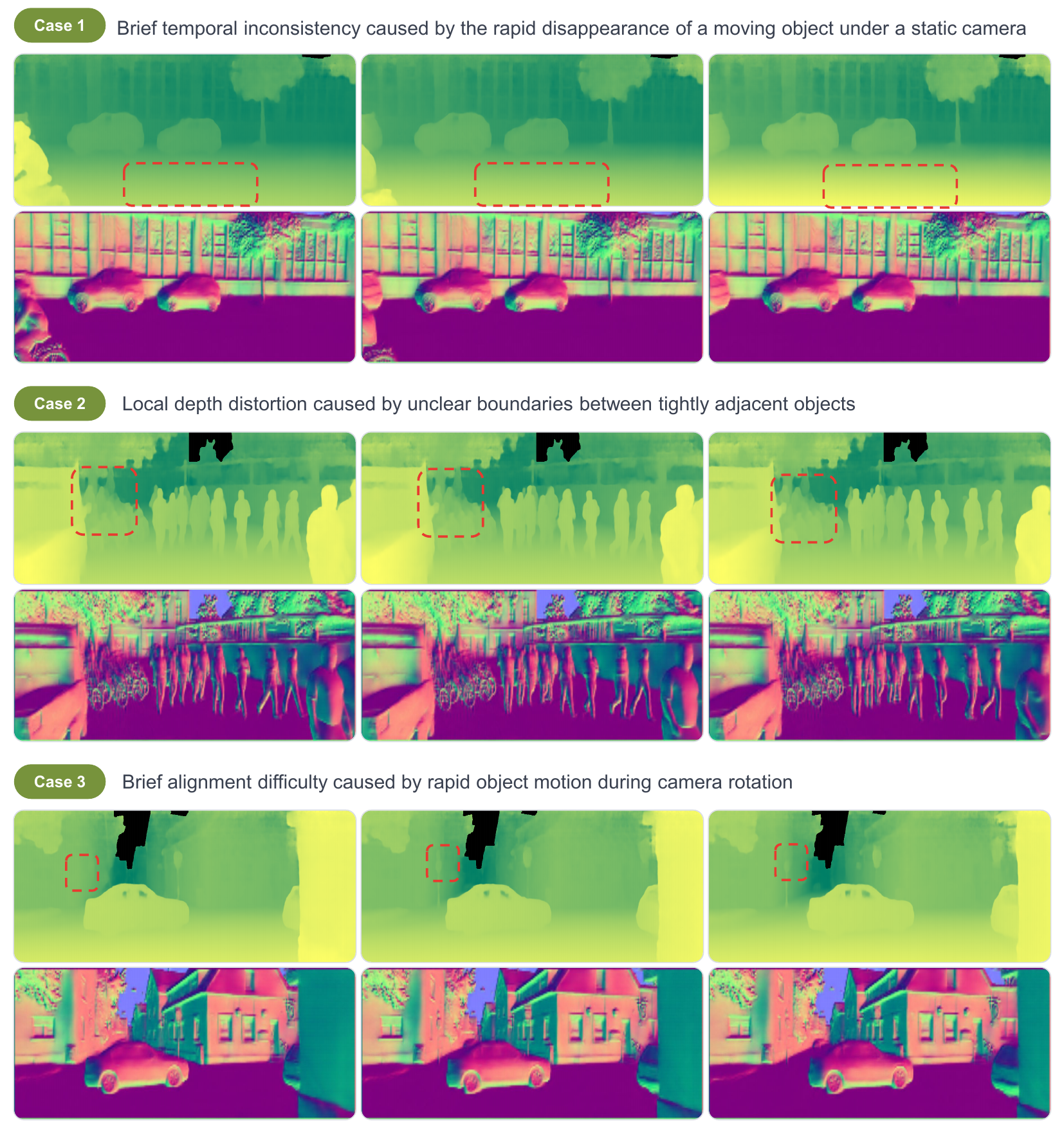}
		\caption{Failure cases. Red dashed boxes highlight failure regions in the predicted depth and normal maps.}
		\label{fig:failure_case}
	\end{center}
\end{figure}

\subsection{Limitations}

Our method depends on the quality of both depth and surface normal estimation, as these cues are used to separate regions and guide temporal learning. As shown in Figure~\ref{fig:failure_case}, challenging motion, occlusion, ambiguous boundaries, or degraded normal estimates may reduce the reliability of the difference mask and lead to local temporal inconsistency. In addition, training requires surface normal supervision, introducing an extra preprocessing step that relies on auxiliary dense depth generation for sparse training data.

Furthermore, since STATIC enforces temporal consistency using only two adjacent frames, it does not explicitly model long-range temporal context. In long videos with substantial scene changes, consistency is propagated only through consecutive frame pairs, which may be less effective than multi-frame approaches using a wider temporal window. Extending STATIC to incorporate longer temporal context while preserving its two-frame efficiency remains an important future direction.

Moreover, since the SNS and MS modules each employ their own attention mechanism on decoding features, two attention operations are required per frame pair. As these operations are applied to spatial feature maps, the computational cost can increase rapidly at higher input resolutions, as reflected in the inference time in Table~\ref{tab:computational_cost}.

Finally, although STATIC is not tied to a specific backbone, its absolute performance may still be influenced by the encoder and depth embedder. As an evaluation limitation, flow-based temporal metrics may inherit flow-estimation errors; however, this shared protocol limitation does not introduce flow dependency into STATIC, and complementary depth metrics and qualitative results are also reported.

\section{Conclusion}

We presented STATIC, a video depth estimation framework that improves temporal consistency by leveraging intrinsic geometric cues from surface normals. Unlike prior approaches that rely on external motion priors or heavy multi-frame fusion, STATIC
uses only two adjacent frames and separately learns geometrically stable and unstable regions. The proposed difference mask identifies regions with noticeable inter-frame geometric variation from surface-normal variation, while the MS and SNS modules improve consistency in the corresponding regions. By integrating these complementary representations, STATIC produces more temporally stable and geometrically coherent depth predictions across consecutive frames. Experimental results on KITTI, NYUv2, and Sintel show that the framework achieves strong temporal consistency and competitive depth accuracy under diverse scene conditions, suggesting that intrinsic geometric cues can effectively guide temporal modeling for video depth estimation.

\newpage
\setstretch{1.2}
 \bibliographystyle{elsarticle-num} 
 \bibliography{cas-refs}

\end{document}